\documentclass[10pt,twocolumn,letterpaper]{article}

\usepackage{cvpr}
\usepackage{times}
\usepackage{epsfig}
\usepackage{graphicx}
\usepackage{amsmath}
\usepackage{amssymb}
\usepackage{mathtools}
\usepackage{xfrac}
\usepackage{pifont}
\usepackage[hang,flushmargin]{footmisc}
\let\norm\undefined 
\DeclarePairedDelimiter\norm{\lVert}{\rVert}
\newcommand{\cmark}{\ding{51}}%
\newcommand{\xmark}{\,}%

\usepackage[pagebackref=true,breaklinks=true,colorlinks,bookmarks=false]{hyperref}

\cvprfinalcopy 

\def\cvprPaperID{5592} 

\ifcvprfinal\fi
\begin{document}

\title{Multi-level Multimodal Common Semantic Space for Image-Phrase Grounding}

\author{Hassan Akbari, ~Svebor Karaman, ~Surabhi Bhargava, ~Brian Chen,\\ Carl Vondrick, and ~Shih-Fu Chang\\\vspace{-0.3cm}
\\
Columbia University, New York, NY, USA\\\vspace{-0.1cm}
{\tt\small
\{ha2436,sk4089,sb4019,bc2754,cv2428,sc250\}@columbia.edu}} 


\maketitle

\begin{abstract}
We address the problem of phrase grounding by learning a multi-level common semantic space shared by the textual and visual modalities. 
We exploit multiple levels of feature maps of a Deep Convolutional Neural Network, 
as well as contextualized word and sentence embeddings extracted from a character-based language model. 
Following dedicated non-linear mappings for visual features at each level, word, and sentence embeddings, we obtain multiple instantiations of our common semantic space in which comparisons between any target text and the visual content 
is performed with cosine similarity.
We guide the model by a multi-level multimodal attention 
mechanism which outputs attended visual features at each level.
The best level is chosen to be compared with text content for maximizing the pertinence scores of image-sentence pairs of the ground truth.
Experiments conducted on three publicly available datasets show 
significant performance gains
 (20\%-60\% relative) over the 
state-of-the-art in phrase localization and set a new performance record on those datasets. 
We provide a detailed ablation study to show the contribution of each element of our approach and release our code on GitHub\footnote{\href{https://github.com/hassanhub/MultiGrounding}{https://github.com/hassanhub/MultiGrounding}}.
\end{abstract}
\vspace{-0.2cm}

\section{Introduction}





\begin{figure}[tb]
    \centering
    \includegraphics[width=\columnwidth]{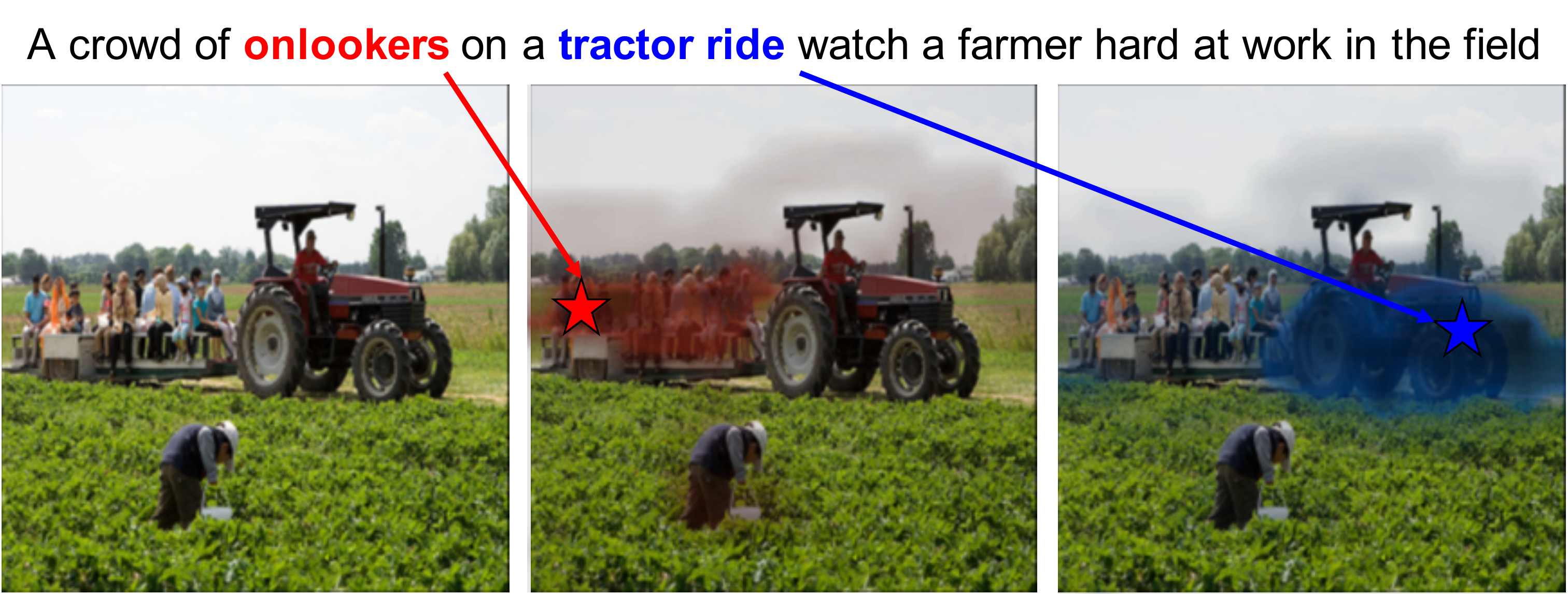}
    \caption{The phrase grounding task in the pointing game setting. Given the sentence on top and the image on the left, the goal is to point (illustrated by the stars here) to the correct location of each natural language query (colored text). Actual example of our method results on Flickr30k.}
    \label{fig:grounding_illustration}
    \vspace{-0.2cm}
\end{figure}

Phrase grounding~\cite{plummer2015flickr30k,mao2016generation} is the task of localizing within an image a given natural language input phrase, as illustrated in Figure~\ref{fig:grounding_illustration}.
This ability to link text and image content is a key component of many visual semantic tasks such as image captioning~\cite{fang2015captions,karpathy2015deep,johnson2016densecap}, visual question answering~\cite{VQA,lu2016hierarchical,xiong2016dynamic,yang2016stacked,fukui2016multimodal}, text-based image retrieval~\cite{gordo2016deep,radenovic2016cnn}, and robotic navigation~\cite{thomason2017guiding}.
It is especially challenging as it requires a good representation of both the visual and textual domain and an effective way of linking them.

On the visual side, most of the works exploit Deep Convolutional Neural Networks but often rely on bounding box proposals~\cite{plummer2015flickr30k,rohrbach2016grounding,hu2016natural} or use a global feature of the image~\cite{fang2015captions}, limiting the localization ability and freedom of the method.
On the textual side, methods rely on a closed vocabulary or try to train their own language model using small image-caption pairs datasets~\cite{javed2018learning,zhang2018top,yeh2017interpretable,engilberge2018finding}.
Finally, the mapping between the two modalities is often performed with a weak linear strategy~\cite{plummer2015flickr30k,xu2018attngan}.
We argue that approaches in the literature have not fully leveraged the potential of the more powerful 
visual
and textual model developed recently, and there is room for developing more sophisticated  
representations and 
mapping approaches. 


In this work, we propose to explicitly learn a non-linear mapping of the visual and textual modalities into a common space, and do so at different granularity for each domain. 
Indeed, different layers of a deep network encode each region of the image with gradually increasing levels of discriminativeness and context awareness, similarly single words and whole sentences contain increasing levels of semantic meaning and context.
This common space mapping is trained 
with weak supervision
and exploited at test-time with a multi-level multimodal attention
mechanism,
where a natural formalism for computing attention heatmaps at each level, attended features and pertinence scoring, enables us to solve the phrase grounding task elegantly and effectively. 
We evaluate our model on three commonly used datasets in the literature of textual grounding and show that it sets a new state-of-the-art performance by a large margin.

Our contributions in this paper are as follows:
\begin{itemize}
\item We learn, with weak-supervision, a non-linear mapping of visual and textual features to a common region-word-sentence semantic space, where comparison between any two semantic representations can be performed with a simple cosine similarity;
\item We propose a multi-level multimodal attention mechanism, producing either word-level or sentence-level attention maps at different semantic levels, enabling us to choose the most representative attended visual feature among different semantic levels; 

\item We set new state-of-the-art performance on three commonly used datasets, and give detailed ablation results showing how each part of our method contributes to the final performance.
\end{itemize}


\section{Related works}


In this section, we give an overview of related works in the literature and discuss how our method differs from them.

\subsection{Grounding natural language in images}



The earliest works on solving  textual grounding~\cite{plummer2015flickr30k,rohrbach2016grounding,hu2016natural} tried to tackle the problem by finding the right bounding box out of a set of proposals, usually obtained from pre-specified models~\cite{zitnick2014edge,uijlings2013selective}.
The ranking of these proposals, for each text query, can be performed using scores estimated from a reconstruction~\cite{rohrbach2016grounding} or sentence generation~\cite{hu2016natural} procedure, or using distances in a common space~\cite{plummer2015flickr30k}.
However, relying on a fixed set of pre-defined concepts and proposals may not be optimal and the quality of the bounding boxes defines an upper bound~\cite{hu2016natural,wang2016structured} of the performance that can be achieved.
Therefore, several methods~\cite{chen2017query,zhaoweakly} have integrated the proposal step in their framework to improve the bounding box quality.
These works often operate in a 
fully supervised setting~\cite{chen2018msrc,yeh2017interpretable,yu2018mattnet,fukui2016multimodal,chen2017query}, where the mapping between sentences and bounding boxes has to be provided at training time which is not always available and is costly to gather.
Furthermore, methods based on bounding boxes often extract features separately for each bounding box~\cite{hu2016natural,chen2018knowledge,wang2016structured}, inducing a high computational cost.

Therefore, some works~\cite{ramanishka2017top,javed2018learning,zhang2018top,xiao2017weakly,yeh2018unsupervised} choose not to rely on bounding boxes and propose to formalize the localization problem as finding a spatial heatmap for the referring expression. 
This setting is mostly weakly-supervised, where at training time only the image and the text (describing either the whole image or some parts of it) are provided but not the corresponding bounding box or segmentation mask for each description.
This is the more general setting we are addressing in this paper.
The top-down approaches~\cite{ramanishka2017top,zhang2018top} and the attention-based approach~\cite{javed2018learning} learn to produce a heatmap for each word of a vocabulary. At test time, all these methods produce the final heatmap by averaging the heatmaps of all the words in the query that exist in the vocabulary.
Several grounding works have also explored the use of additional knowledge, such as image~\cite{wang2016structured} and linguistic~\cite{xiao2017weakly,plummer2017phrase} structures, phrase context~\cite{chen2018msrc} and exploiting pre-trained visual models predictions~\cite{chen2018knowledge,yeh2018unsupervised}.

In contrast to many works in the literature, we don't use a pre-defined set of image concepts or  words in our method. 
We instead rely on visual feature maps and a character-based language model with contextualized embeddings which could handle any unseen word considering the context in the sentence. 

\subsection{Mapping to common space}
It is a common approach to extract visual and language features independently and fuse them before the prediction~\cite{engilberge2018finding, chen2018knowledge,chen2017query}. 
Current works usually apply a multi-layer perceptron (MLP)~\cite{chen2017query,chen2018knowledge}, element-wise multiplication~\cite{hendricks2018grounding}, or cosine similarity~\cite{engilberge2018finding} to combine representations from different modalities.
Other methods have used the Canonical Correlation Analysis (CCA)~\cite{plummer2017phrase, plummer2015flickr30k}, which finds linear projections that maximize the correlation between projected vectors from the two views of heterogeneous data. ~\cite{fukui2016multimodal} introduced the Multimodal Compact Bilinear (MCB) pooling method that uses a compressed feature from the outer product of two vectors of visual and language features 
to fuse them.
Attention methods~\cite{xu2018attngan,nguyen2018improved} can also measure the matching of an image-sentence feature pair. 

We use non-linear mappings of both visual features (in multiple semantic levels) and textual embeddings (both contextualized word and sentence embeddings) separately and use multi-level attention with multimodal loss to learn those mapping weights.


\subsection{Attention mechanisms}
Attention has proved its effectiveness in many visual and language tasks~\cite{khademi2018image,anderson2018bottom,chen2017sca,yang2016stacked,xu2015show}, it is designed to capture a better representation of image-sentence pairs based on their interactions.
The Accumulated Attention method~\cite{deng2018visual} propose to estimate attention on sentences, objects and visual feature maps in an iterative fashion, where at each iteration the attention of the other two modalities is exploited as guidance.
A dense co-attention mechanism is explored in~\cite{nguyen2018improved} to solve the Visual Question Answering task by using a fully symmetric architecture between visual and language representations. 
In their attention mechanism, they add a dummy location in attention map when no region or word the model should attend along with a softmax. 
In AttnGAN~\cite{xu2018attngan}, a deep attention multimodal similarity model is proposed to compute a fine-grained image-text matching loss.

In contrast to these works, we remove the softmax on top of the attention maps to let the model decide which word-region could be related to each other by the guide of the multimodal loss. 
Since we map the visual features to a multi-level visual representation, we give the model the freedom to choose any location at any level for either sentence or word. 
In other words, each word or sentence can choose which level of representation (and which region in that representation) to attend to. 
We directly calculate the attention map by cosine similarity in our common semantic space.
We show that this approach significantly outperforms all the state of the art approaches on three commonly used datasets and set a new state of the art performance.

\begin{figure}
    \centering
    \includegraphics[width=0.85\columnwidth]{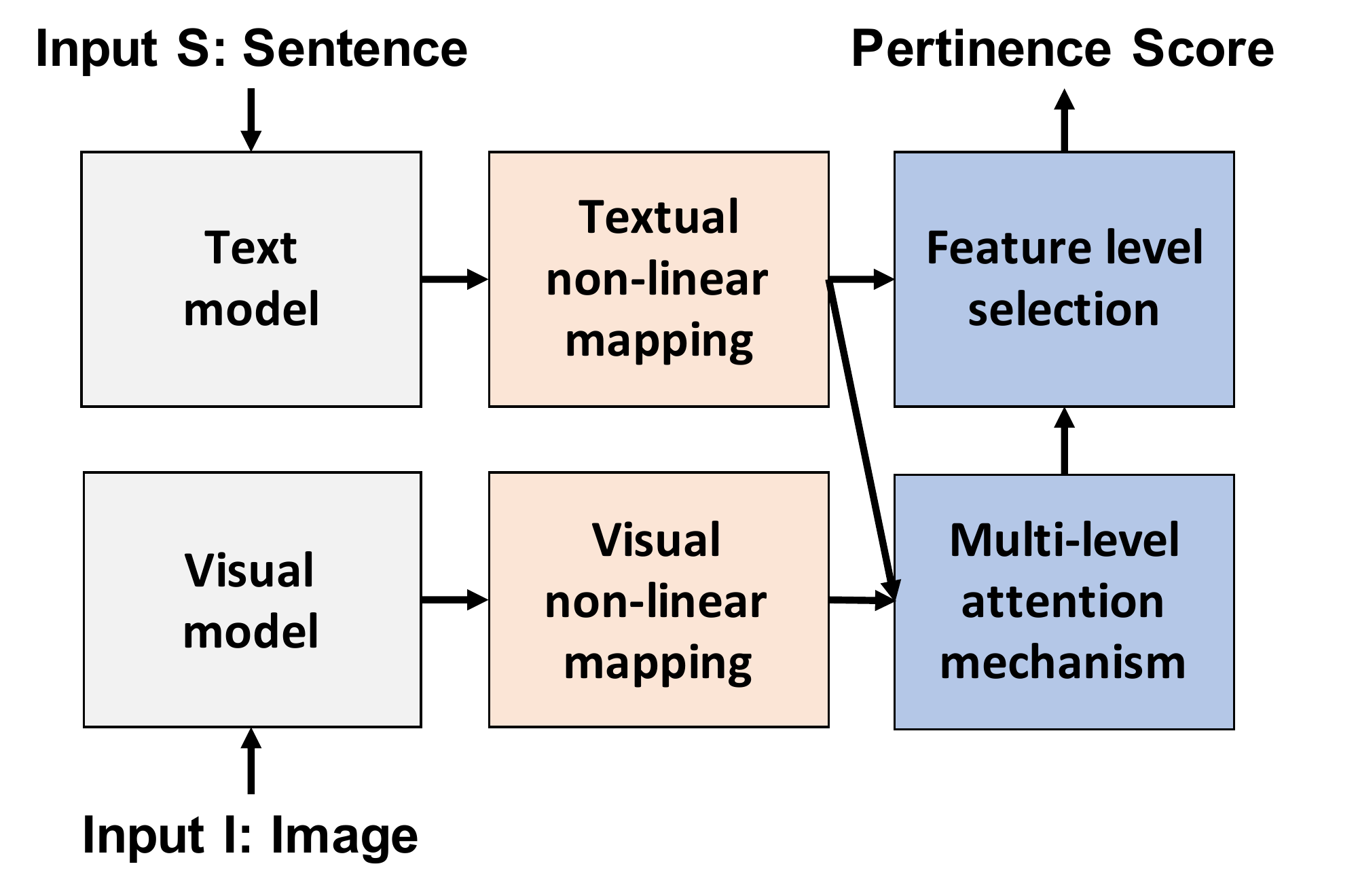}
    \caption{Overview of our method: the textual input is processed with a pre-trained text model followed by a non-linear mapping to the common semantic space.
    Similarly for the image input, we use a pre-trained visual model to extract visual features maps at multiple levels and learn a non-linear mapping for each of them to the common semantic space.
    A multi-level attention mechanism followed by a feature level selection produces the pertinence score between the image and the sentence.
    We train our model
    using only the weak supervision of image-sentence pairs.
    }
    \label{fig:overview}
\end{figure}

\section{Method}

In this section, we describe our method (illustrated in Figure~\ref{fig:overview}) for addressing the textual grounding task and elaborate on each part with details. 
In Section~\ref{sec:feature_ext_map}, we explain how we extract multi-level visual features from an image and word/sentence embeddings from the text, and then describe how we map them to a common space. 
In Section~\ref{sec:mlmm_attention} we describe how we calculate multi-level multimodal attention map and attended visual feature for each word/sentence. 
Then, in Section~\ref{sec:feature_level} we describe how we choose the most representative visual feature level for the given text. 
Finally, in Section~\ref{sec:mm_loss} we define a multimodal loss to train the whole model with weak supervision.

\subsection{Feature Extraction and Common Space}
\label{sec:feature_ext_map}
\begin{figure*}[tb]
    \centering
    \includegraphics[width=1.02\columnwidth]{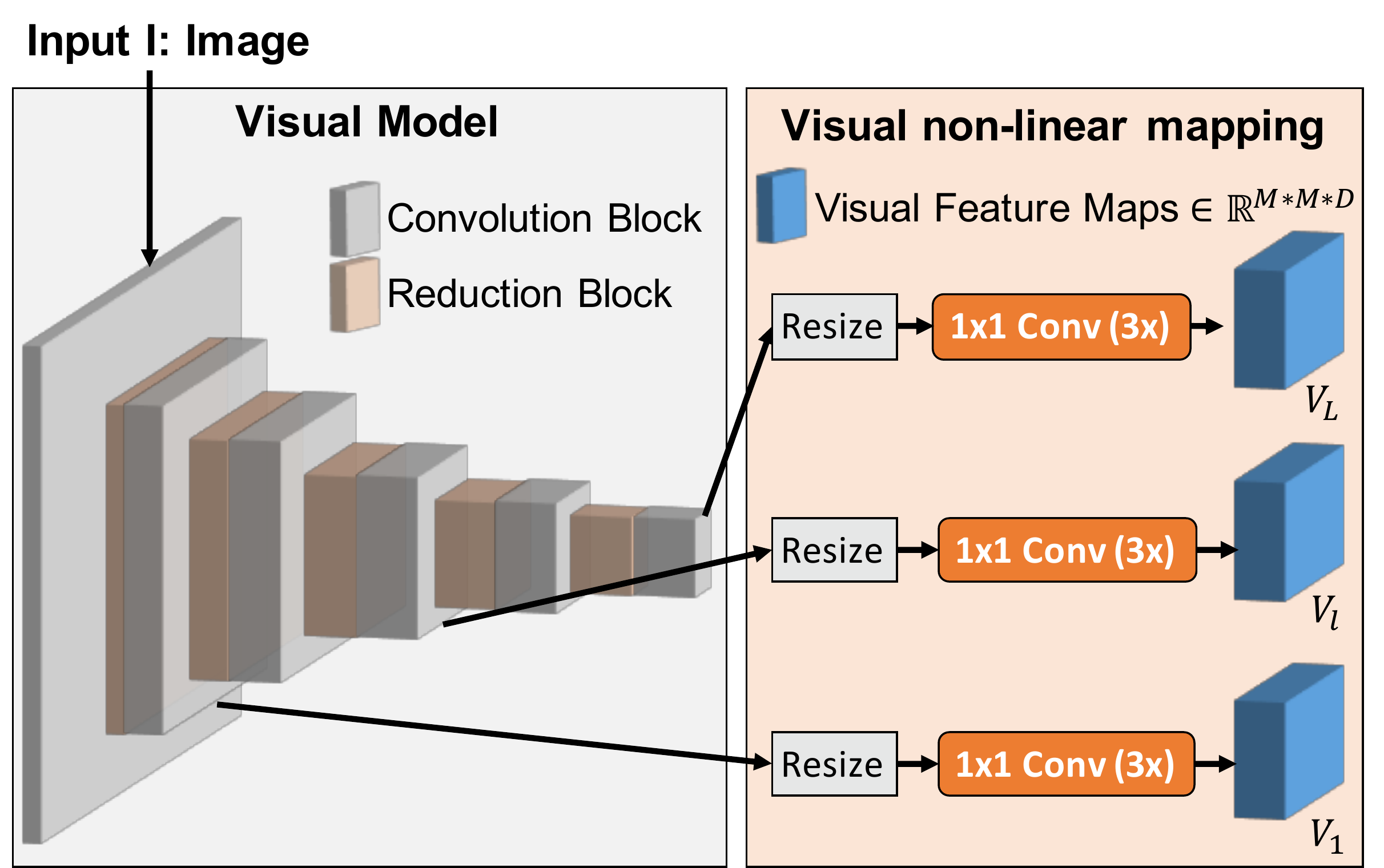}
    \hfill
    \includegraphics[width=0.94\columnwidth]{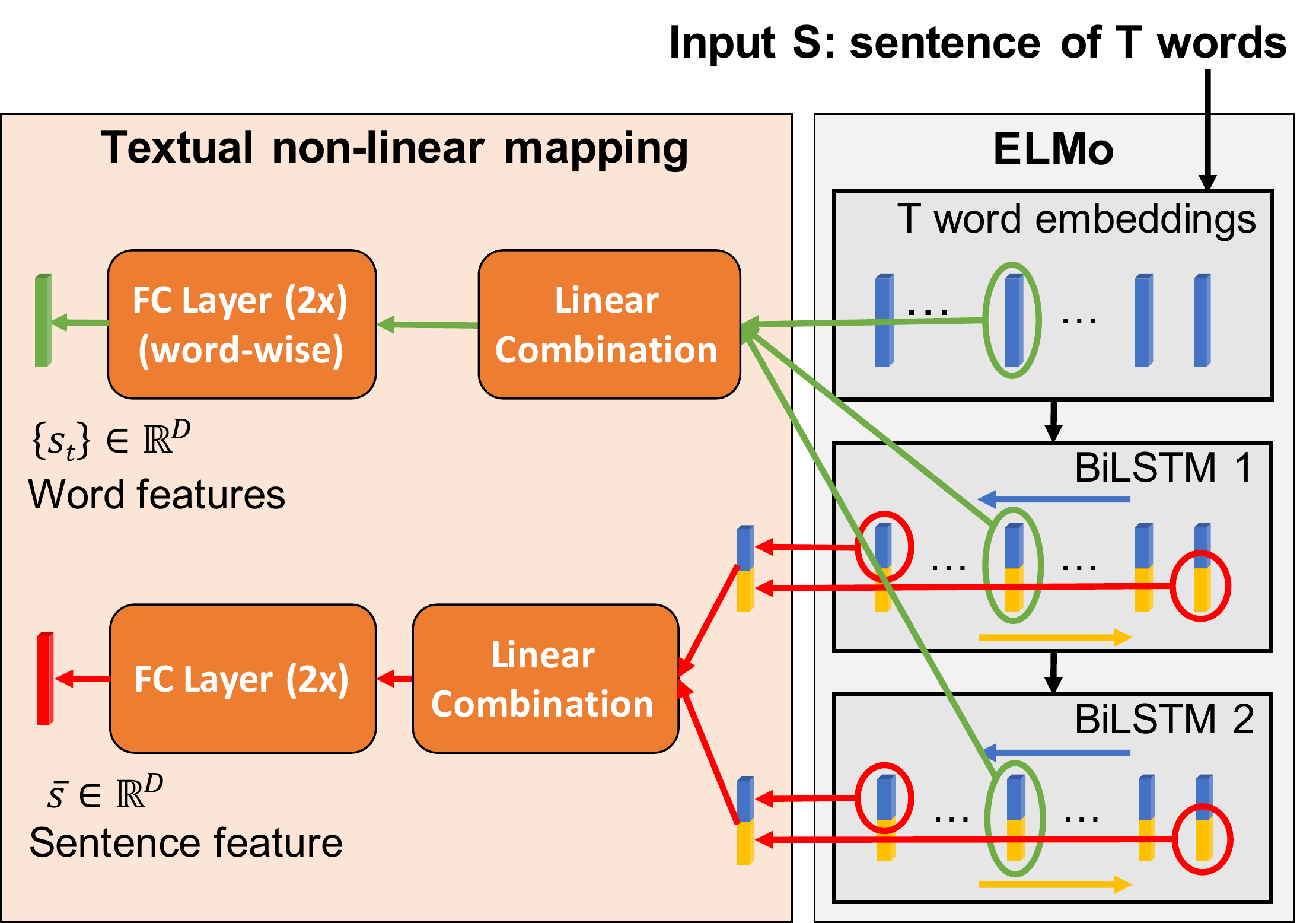}
    \caption{Left: we choose feature maps of different convolutional blocks of a CNN model, resize them to the same spatial dimensions using bi-linear interpolation, and map them to feature maps of the same size.
    Right: word and sentences embedding to the common space from the pre-trained ELMo~\cite{peters2018deep} model. The green pathway is for word embedding, the red pathway for sentence embedding. 
    All the orange boxes ($1 \times 1$ convolutional layers of the visual mapping, linear combination and the two sets of fully connected layers of the textual mapping) are the trainable parameters of our projection to the common space.}
    \label{fig:mapping}
\end{figure*}

\paragraph{Visual Feature Extraction}\hspace{-0.9em}:
In contrast to many vision tasks
where the last layer of a pre-trained CNN is being used as visual representation of an image, we use feature maps from different layers and map them separately to a common space to obtain a multi-level set of feature maps to be compared with text. Intuitively, using different levels of visual representations would be necessary for covering a wide range of visual concepts and patterns \cite{lin2017feature,yosinski2015understanding,zeiler2014visualizing}.
Thus, we extract $L=4$ sets of feature maps from $L$ different levels of a visual network, upsample them by a bi-linear interpolation\footnote{as transposed convolution produces checkerboard artifacts~\cite{odena2016deconvolution}} to 
a fixed resolution $M \times M$ for all the $L$ levels, 
and then apply 3 layers of 1x1 convolution (with LeakyRelu~\cite{maas2013rectifier}) with $D$ filters to map them into equal-sized feature maps. 
Finally, we stack these feature maps and space-flatten them to have an overall image representation tensor $V \in \mathbb{R}^{N \times L \times D}$, with $N = M \times M$. 
This tensor is finally normalized by 
the $l_2$-norm 
of its last dimension. An overview of the feature extraction and common space mapping for image can be seen in the left part of
Figure~\ref{fig:mapping}.

In this work, we use VGG~\cite{Simonyan14c} as a baseline for fair comparison with other works~\cite{fang2015captions,xiao2017weakly,javed2018learning}, and the state of the art CNN, PNASNet-5~\cite{liu2017progressive}, 
to study the ability of our model to exploit this more powerful visual model. 



\paragraph{Textual Feature Extraction}\hspace{-0.9em}:
State-of-the-art works in grounding use a variety of approaches for textual feature extraction. 
Some use pre-trained LSTM or BiLSTMs on big datasets (e.g. Google 1 Billion~\cite{chelba2014one}) based on either word2vec~\cite{mikolov2013linguistic} or GloVe~\cite{pennington2014glove} representations. 
Some train BiLSTM solely on image-caption datasets (mostly MSCOCO) and argue it is necessary to train them from scratch to distinguish between visual concepts which may not be distinguishable in language (e.g. red and green are different in vision but similar in language as they are both colors) \cite{nguyen2018improved,xu2018attngan,javed2018learning,xiao2017weakly,engilberge2018finding,hendricks2018grounding,zhaoweakly,plummer2015flickr30k,yu2018mattnet,deng2018visual}.
These works either use the recurrent network outputs at each state as word-level representations or their last output (on each direction for BiLSTM) as sentence-level or a combination of both.

In this paper, however we use ELMo~\cite{peters2018deep}, a 3-layer network pre-trained on 5.5B tokens 
which calculates word representations on the fly 
(based on CNN on characters, similar to \cite{jozefowicz2016exploring,zhang2015character})
and then feed them to 2 layers of BiLSTMs which produce contextualized representations. 
Thus, for a given sentence the model outputs three representations for each token (splitted by white space). 
We take a linear combination of the three representations 
and feed them to 2 fully connected layers (with shared weights among words), each with $D$ nodes with LeakyRelu as non-linearity between each layer, to obtain each word representation $\mathbf{s}_t$
(green pathway in the right part of
Figure~\ref{fig:mapping}).
The resulting word-based text representation for an entire sentence would be a tensor $\mathbf{S} \in \mathbb{R}^{T \times D}$ built from the stacking of each word representation $\mathbf{s}_t$. The sentence-level text representation is calculated by concatenation of last output of the BiLSTMs at each direction. 
Similarly, we apply a linear combination on the two sentence-level representations 
and map it to the common space by feeding it to 2 fully connected layers of $D$ nodes, producing the sentence representation $\overline{\mathbf{s}}$ (red pathway in the right part of Figure~\ref{fig:mapping}). 
The word tensor and the sentence vector are normalized by their last dimension $l_2$-norm before being fed to the multimodal attention block. 



\subsection{Multi-Level Multimodal Attention Mechanism}
\label{sec:mlmm_attention}
Given the image and sentence, our task is to estimate the correspondences between spatial regions ($n$) in the image at different levels ($l$), and words in the sentence at different positions ($t$).
We seek to estimate a correspondence measure, $H_{n,t,l}$, between each word and each region at each level. We define this correspondence by the cosine similarity between word and image region representations at different levels in common space:
\begin{equation}
    \label{eq:heatmap_multi}
    H_{n,t,l} = \max(0, \langle \mathbf{s}_{t}, \mathbf{v}_{n,l}\rangle).
\end{equation}
Thus, $\mathbf{H} \in \mathbb{R}^{N \times T \times L}$ represents a multi-level multi-modal attention map which could be simply used for calculating either visual or textual attended representation. 
We apply ReLU to the attention map to zero-out dissimilar word-visual region pairs, and simply avoid applying softmax on any dimension of the heatmap tensor. 
Note that this choice is very different in spirit from the 
commonly used approach of applying softmax to attention maps~\cite{xu2015show,xu2016ask,deng2018visual,nguyen2018improved,javed2018learning,xu2018attngan,ramanishka2017top}.
Indeed for irrelevant image-sentence pairs, the attention maps would be almost all zeros while the softmax process would always force attention to be a distribution over the image/words summing to $1$. 
Furthermore, a group of words shaping a phrase could have the same attention area which is again hard to achieve considering the competition among regions/words in the case of applying softmax on the heatmap. We will analyze the influence of this choice experimentally in our ablation study.

\begin{figure}[tb]
    \centering
    \includegraphics[width=\columnwidth]{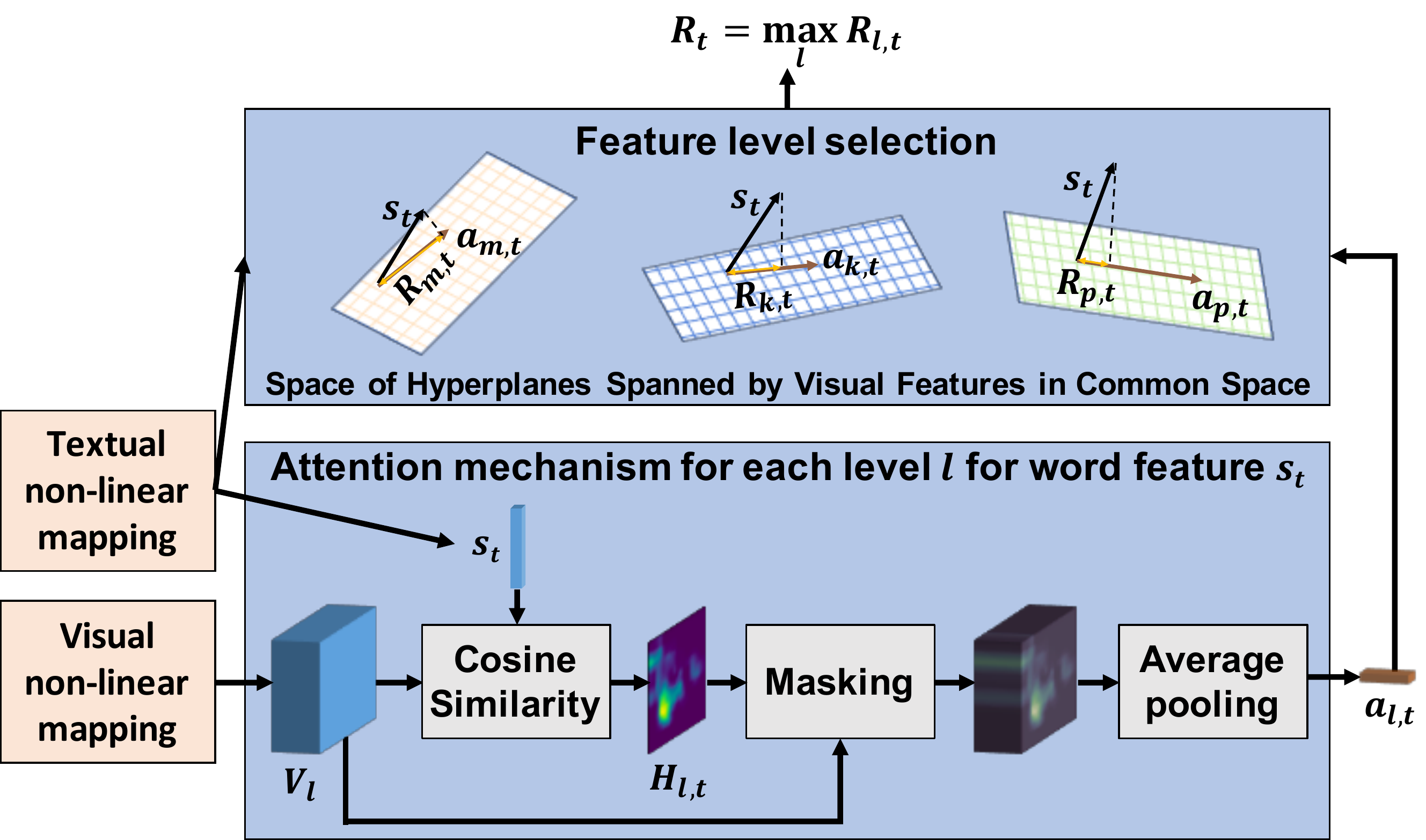}
    \caption{For each word feature $s_t$, we compute an attention map $H_{l,t}$ and an attended visual feature $a_{l,t}$ at each level $l$. 
    We choose the level that maximizes similarity between the attended visual feature and the textual feature in the common space to produce the pertinence score $R_t$. 
    This is equivalent to finding the hyperplane (spanned by each level visual feature vectors in the common space) that best matches the textual feature.}
    \label{fig:attention}
\end{figure}

Given the heatmap tensor, we calculate the attended visual feature for the $l$-th level and $t$-th word as
\begin{equation}
    \mathbf{a}_{t,l} = \frac{\sum_{n=1}^{N}H_{t,n,l}\mathbf{v}_{n,l}}{\norm[\Big]{\sum_{n=1}^{N}H_{t,n,l}\mathbf{v}_{n,l}}_{2}},
\end{equation}

\noindent which is basically a weighted average over the visual representations of the $l$-th level with the attention heatmap values as weights. 
In other words, $\mathbf{a}_{t,l}$ is a vector in the hyperplane spanned by a subset of visual representations in the common space, this subset being selected based on the heatmap tensor.
An overview of our multi-level multimodal attention mechanism for calculating attended visual feature can be seen in 
Figure~\ref{fig:attention}. 
In the sequel, we describe how we use this attended feature to choose the most representative hyperplane, and calculate a multimodal loss to be minimized by weak supervision of image-sentence relevance labels.


\subsection{Feature Level Selection}
\label{sec:feature_level}
Once we find the attended visual feature, we calculate the word-image pertinence score at level $l$ using cosine similarity for each word and the attended visual feature as
\begin{equation}
    R_{t,l} = \langle \mathbf{a}_{t,l}, \mathbf{s}_{t}\rangle.
\end{equation}
Intuitively, each visual feature map level could carry different semantic information, thus for each word we propose to apply a hard level-attention to get the score from the level contributing the most as
\begin{equation}
\label{word_pertinence}
R_{t} = \max_{l}{R_{t,l}}.
\end{equation}
This procedure can be seen as finding projection of the textual embeddings on hyperplanes spanned by visual features from different levels and choosing the one that maximizes this projection. Intuitively, that chosen hyperplane can be a better representation for visual feature space attended by word $t$. 
This can be seen in the top central part of Figure~\ref{fig:overview}, 
where selecting the maximum pertinence score over levels is equivalent to selecting the hyperplane with the smallest angle with the $t$-th word representation (or the highest similarity between attended visual feature and textual feature). Thus, selecting the most representative hyperplane (or visual feature level).


Once we find the best word-image pertinence score, similar to~\cite{xu2018attngan} and inspired by the minimum classification error~\cite{juang1997minimum}, we find the overall (word-based) sentence-image pertinence score as follows:
\begin{equation}
\label{w_matching}
    R_w(S,I) = \log\bigg( {\Big( \sum_{t=0}^{T-1}\exp{(\gamma_1R_t)}  \Big)}^{\frac{1}{\gamma_1}} \bigg).
\end{equation}

Similarly, for the sentence we can repeat the same procedure (except that we no more need Eq.~(\ref{w_matching})) for finding the attention map, attended visual feature and sentence-image pertinence score as follows, respectively: 
\begin{subequations}
\setlength{\jot}{7pt}
\begin{align}
    &H_{n,l}^{s} = \max(0,\langle \bar{\mathbf{s}}, \mathbf{v}_{n,l}\rangle)\label{rs} \\
    &\mathbf{a}_{l}^{s} = \sum_{n=1}^{N}H_{n,l}^{s}\mathbf{v}_{n,l} \\
    &R_{s,l} = \langle \mathbf{a}_{l}^{s},  \bar{\mathbf{s}}\rangle \\
    &R_s(S,I) = \max_{l}{ R_{s,l}}
\end{align}
\end{subequations}

\subsection{Multimodal Loss}
\label{sec:mm_loss}
In this paper, we only use a weak supervision in the form of binary image-caption relevance. 
Thus, similar to
~\cite{fang2015captions,huang2013learning,xu2018attngan}
we train the network on a batch of image-caption pairs, $\{(S_b,I_b)\}_{b=1}^{B}$ and force it to have high sentence-image pertinence score for related pairs and low score for unrelated pairs. 
Thus, considering a pertinence score $R_x$ (either $R_w$ or $R_s$), we calculate the posterior probability of the sentence $S_b$ being matched with image $I_b$ by applying competition among all sentences in the batch using:
\begin{equation}\label{prob1}
    P_x(S_b|I_b) = \frac{\exp(\gamma_2R_x(S_b,I_b))}{\sum_{b'}^{B}{\exp(\gamma_2R_x(S_{b'},I_b))}}
\end{equation}
Similarly, the posterior probability of $I_b$ being matched with $S_b$ could be calculated using:
\begin{equation}\label{prob2}
    P_x(I_b|S_b) = \frac{\exp(\gamma_2R_x(S_b,I_b))}{\sum_{b'}^{B}{\exp(\gamma_2R_x(S_b,I_{b'}))}}
\end{equation}
Then, similarly to~\cite{fang2015captions, xu2018attngan}, we can define the loss using the negative log posterior probability over relevant image-sentence pairs as follows:
\begin{equation}
    L^{x} = -\sum_{b=1}^{B}\Big(\log{P_x(S_b|I_b)}+\log{P_x(I_b|S_b)}\Big)
\end{equation}

As we want to train a common semantic space for both words and sentences, we combine the loss $L^w$ (that can be computed based on the word relevance $R_w$) and the sentence loss $L^s$ (obtained using $R_s$) to define our final loss $L$ as

\begin{equation}
    L = L^{w} + L^{s}.
\end{equation}
This loss is minimized over a batch of $B$ images along with their related sentences.
We found in preliminary experiments on held-out validation data, that the values $\gamma_1=5$, $\gamma_2=10$ work well and we keep them fixed for our experiments.
In the next section, we will evaluate our proposed model on different datasets and will have an ablation study to show the reason for our choices in our model.

\section{Experiments}
In this section, we first present the datasets used in our experimental setup. We then evaluate our approach comparing with the state-of-the-art, and further present ablation studies showing the influence of each step of our method.

\subsection{Datasets}
\label{datasets}

\paragraph{MSCOCO 2014}\hspace{-0.6em}\cite{lin2014microsoft} consists of 82,783 training images and 40,504 validation images. Each image is associated with five captions describing the image. 
We use the train split of this dataset for training our model.

\vspace{-0.5em}
\paragraph{Flickr30k Entities}\hspace{-0.6em}\cite{plummer2015flickr30k} contains 224k phrases describing localized bounding boxes in $\sim$31k images each described by 5 captions. Images and captions come from Flickr30k~\cite{young2014image}. We use 1k images from the test split of this dataset for evaluation.


\vspace{-0.5em}
\paragraph{VisualGenome}
\hspace{-0.6em}\cite{krishna2017visual} contains 77,398 images in the training set, and a validation and test set of 5000 images each.
Each image consists of multiple bounding box annotations and a region description associated with each bounding box. We use the train split of this dataset to train our models and use its test split for evaluation.

\vspace{-0.5em}
\paragraph{ReferIt}\hspace{-0.6em}consists of 20,000 images from the IAPR TC-12 dataset~\cite{grubinger2006iapr} along with 99,535 segmented image regions from the SAIAPR-12 dataset~\cite{chen2017query}. 
Images are associated with descriptions for the entire image as well as localized image regions collected in a two-player game~\cite{kazemzadeh2014referitgame} providing approximately 130k isolated entity descriptions. 
In our work, we only use the unique descriptions associated with each region. We use a split similar to~\cite{hu2016natural} which contains 9k training, 1k validation, and 10k test images. We use the test split of this dataset to evaluate our models.

\begin{table}[tb]
\centering
\resizebox{\columnwidth}{!}{%
\begin{tabular}{|c|c|c|c|c|c|}
\cline{4-6}
\multicolumn{3}{}{} & \multicolumn{3}{|c|}{Test Accuracy} \\
\hline
Method & Settings & Training & VG & Flickr30k & ReferIt \\
\hline
Baseline & Random & - & 11.15 & 27.24 & 24.30 \\ 
Baseline & Center & - & 20.55 & 49.20 & 30.40 \\
\hline
TD~\cite{zhang2018top} & Inception-2 & VG & 19.31 & 42.40 & 31.97 \\
SSS~\cite{javed2018learning} & VGG & VG & 30.03 & 49.10 & 39.98 \\
Ours & BiLSTM+VGG & VG & 50.18 & 57.91 & \textbf{62.76}  \\
Ours & ELMo+VGG & VG & 48.76 & 60.08 & 60.01  \\
Ours & ELMo+PNASNet & VG & \textbf{55.16} & \textbf{67.60} & 61.89 \\
\hline
CGVS~\cite{ramanishka2017top} & Inception-3 & MSR-VTT & - & 50.10 & -  \\
\hline
FCVC~\cite{fang2015captions} & VGG & MSCOCO & 14.03 & 29.03 & 33.52 \\
VGLS~\cite{xiao2017weakly} & VGG & MSCOCO & 24.40 & - & - \\
Ours & BiLSTM+VGG & MSCOCO & 46.99 & 53.29 & 47.89  \\
Ours & ELMo+VGG & MSCOCO & 47.94 & 61.66 & 47.52  \\
Ours & ELMo+PNASNet & MSCOCO & \textbf{52.33} & \textbf{69.19} & \textbf{48.42}  \\
\hline
\end{tabular}
}
\caption{Phrase localization accuracy (pointing game) on Flickr30k, ReferIt and VisualGenome (VG) compared to state of the art methods.}
\label{tab:res-pointing-3datasets}
\vspace{-0.25cm}
\end{table}

\subsection{Experimental Setup}
\label{sec:experimental_setup}
We use a batch size of $B=32$, where for a batch of image-caption pairs each image (caption) is only related to one caption (image). 
Image-caption pairs are sampled randomly with a uniform distribution. 
We train the network for 20 epochs with the Adam optimizer~\cite{kingma2014adam} with $lr=0.001$ where the learning rate is divided by 2 once at the 10-th epoch and again at the 15-th epoch. 
We use $D=1024$ for common space mapping dimension and $\alpha=0.25$ for LeakyReLU in the non-linear mappings. 
We regularize weights of the mappings with $l_2$ regularization with reg$\_$value $=0.0005$. 
For VGG, we take outputs from \{conv4\_1, conv4\_3, conv5\_1, conv5\_3\} and map to semantic feature maps with dimension $18\times18\times1024$, and for PNASNet we take outputs from \{Cell 5, Cell 7, Cell 9, Cell 11\} and map to features with dimension $19\times19\times1024$.
Both visual and textual networks weights are fixed during training and only common space mapping weights are trainable. 
In the ablation study, we use 10 epochs without dividing learning rate, while the rest of settings remain the same.
We follow the same procedure as in~\cite{javed2018learning,johnson2016densecap,plummer2015flickr30k,xiao2017weakly} for cleaning and pre-processing the datasets and use the same train/test splits for fair comparison in our evaluations.

\subsection{Phrase Localization Evaluation}
\label{sec:eval}

\begin{table}[tb]
\centering
\resizebox{\columnwidth}{!}{%
\begin{tabular}{|c|c|c|c|c|c|c|}
\cline{2-7}
 \multicolumn{1}{}{} & \multicolumn{3}{|c|}{pointing game accuracy}  & \multicolumn{3}{|c|}{attention correctness} \\
\hline
 & \cite{ramanishka2017top} & Ours & Ours & \cite{ramanishka2017top} & Ours & Ours \\
Class & Inc.3 & VGG & PNAS & Inc.3 & VGG & PNAS \\
\hline
bodyparts & 0.194 & 0.408 & \bf{0.449} & 0.155 & 0.299 & \bf{0.373} \\
\hline
animals & 0.690 & 0.867 & \bf{0.876} & 0.657 & 0.701 & \bf{0.826} \\
\hline
people & 0.601 & 0.673 & \bf{0.756} & 0.570 & 0.562 & \bf{0.724} \\
\hline
instrument & 0.458 & 0.286 & \bf{0.575} & 0.502 & 0.297 & \bf{0.555} \\
\hline
vehicles & 0.645 & 0.781 & \bf{0.838} & 0.615 & 0.554 & \bf{0.738} \\
\hline
scene & 0.667 & \bf{0.685} & 0.682 & 0.582 & 0.596 & \bf{0.639} \\
\hline
other & 0.427 & 0.502 & \bf{0.598} & 0.348 & 0.424 & \bf{0.535} \\
\hline
clothing & 0.360 & 0.472 & \bf{0.583} & 0.345 & 0.330 & \bf{0.473} \\
\hline
\hline
average & 0.501 & 0.617 & \bf{0.692} & 0.473 & 0.508 & \bf{0.639} \\
\hline
\end{tabular}
}
\caption{Category-wise pointing game accuracy and attention correctness on Flickr30k Entities.}
\label{tab:results_category}
\end{table}


\begin{figure*}[ht]
    \centering
    \includegraphics[width=\textwidth]{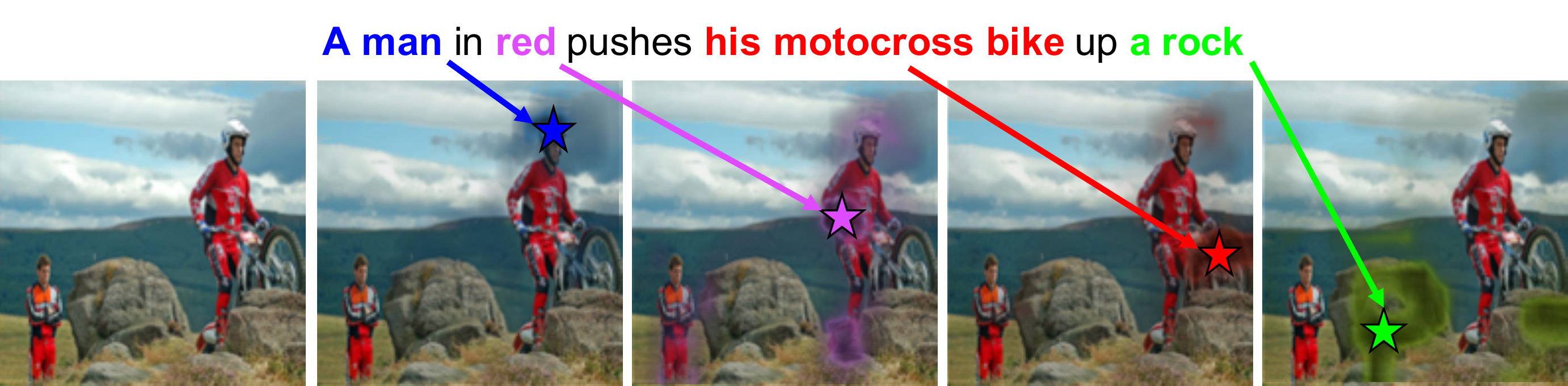}\\
    \caption{Image-sentence pair from Flickr30k with four queries (colored text) and corresponding heatmaps and selected max value (stars).}
    \label{fig:qualitative_flickr30k_big}
    \vspace{-0.15cm}
\end{figure*}

As stated in Section~\ref{datasets}, we train our model on the train split of MSCOCO and Visual Genome (VG), and evaluate it on the test splits of Flickr30k, ReferIt, and VG.
In test time, for Flickr30k we feed a complete sentence to the model and take weighted average of attention heatmaps of words for each query with the word-image pertinence scores from Eq.~(\ref{word_pertinence}) as weights. 
For ReferIt and Visual Genome, we treat each query as a single sentence and take its sentence-level attention heatmap as the final query pointing heatmap. 
Once the pointing heatmaps are calculated, we find the max location (as pointing location for the given query) and evaluate the model by the pointing game accuracy: $\frac{\#hit}{\#hit+\#miss}$.

Pointing game accuracy results can be found in Table~\ref{tab:res-pointing-3datasets} for Flickr30k, ReferIt and Visual Genome datasets.
The results show that our method significantly outperforms all state-of-the-art methods in all conditions and all datasets. 
For fair comparison with~\cite{javed2018learning,fang2015captions,xiao2017weakly}, we used a VGG16 visual model and replaced the pre-trained BiLSTM layers of ELMo with a single trainable BiLSTM. 
This model (BiLSTM+VGG) still gives a pointing game accuracy absolute improvement of $20.15\%$ for VisualGenome, $7.81\%$ for Flickr30k, and $23.28\%$ for ReferIt, while giving relative improvement of $67.09\%$, $15.59\%$, and $56.98\%$, respectively. 
Results with the more recent PNASNet model are even  better, especially for Flickr30k and VisualGenome. 

To get a deeper understanding of our model, 
we first report in Table~\ref{tab:results_category} category-wise pointing game accuracy and attention correctness~\cite{liu2017attention} (percentage of the heatmap falling into the ground truth bounding box) and compare with the state-of-the-art method~\cite{ramanishka2017top}
on Flickr30k.
We observe that our method obtains a higher performance on almost all categories even when VGG16 is used as the visual backbone. The model based on PNASNet consistently outperforms the state-of-the-art on all categories on both metrics. 
We further perform a test on level selection rate for different types of queries and report them in Table \ref{tab:level_selection}. 
It shows that the 3rd level
dominates the selection while the 4th level is also important for several categories such as scene and animals. The 1st level is exploited mostly for the animals and people categories.
The full sentence selection relies mostly on the 3rd level as well, while for some sentences the 4th model has been selected.
This demonstrates the power of the proposed method in
selecting the right level of representation.

\begin{table}[tb]
\centering
\resizebox{\columnwidth}{!}{%
\begin{tabular}{|c|c|c|c|c|c|c|c|c||c||c|}
\cline{2-11}
\multicolumn{1}{}{} & \multicolumn{10}{|c|}{Selection Rate ($\%$)} \\
\hline
 \shortstack{\\Level / \\PNASNet \\ Layers} & \rotatebox[origin=c]{270}{bodyparts} & \rotatebox[origin=c]{270}{animals} & \rotatebox[origin=c]{270}{people} & \rotatebox[origin=c]{270}{instrument} & \rotatebox[origin=c]{270}{vehicles} & \rotatebox[origin=c]{270}{scene} & \rotatebox[origin=c]{270}{other} & \rotatebox[origin=c]{270}{clothing} & \rotatebox[origin=c]{270}{average} & \rotatebox[origin=c]{270}{sentence} \\
\hline
1 / Cell 5 & 2.6 & 10.4 & 7.5 & 0.9 & 2.0 & 5.4 & 5.4 & 5.3 & 6.3 & 0.7\\
\hline
2 / Cell 7 & 0.1 & 2.0 & 4.2 & 0.0 & 1.7 & 2.5 & 0.9 & 0.3 & 2.5 & 0.05\\
\hline
3 / Cell 9 & 85.9 & 48.4 & 64.6 & 88.6 & 68.3 & 49.5 & 70.9 & 86.1 & 66.5 & 86.51\\
\hline
4 / Cell 11 & 11.4 & 39.2 & 23.7 & 10.5 & 27.9 & 42.6 & 22.8 & 8.3 & 24.7 & 12.7\\
\hline
\end{tabular}
}
\caption{Level selection rate for different layers of PNASNet on different categories in Flickr30k}
\label{tab:level_selection}
\vspace{-0.15cm}
\end{table}

\subsection{Ablation Study}

In this section, we trained on MSCOCO multiple configurations of our approach, with a PNASNet visual model, to better understand which aspects of our method affects positively or negatively the performance. 
We report evaluation results on Flickr30k in Table~\ref{tab:ablation}. Results are sorted by performance to show the most successful combinations.


\begin{table}[tb]
    \centering
\resizebox{0.85\columnwidth}{!}{%
\begin{tabular}{|c|c|c|c|c|c|c|c|}
\cline{2-8}
 \multicolumn{1}{c|}{} & SA & ELMo & NLT & NLV & WL & SL & Acc.\\
\hline
1 & \xmark & \cmark & \cmark & \cmark & ML & ML & 67.73 \\ 
\hline
2 & \xmark & \cmark & \cmark & \cmark & M & L & 62.67 \\ 
\hline
3 & \xmark & \xmark & \cmark & \cmark & ML & ML & 61.13 \\ 
\hline
4 & \xmark & \cmark & \xmark & \cmark & M & L & 58.40 \\ 
\hline
5 & \xmark & \cmark & \cmark & \xmark & M & L & 56.92 \\ 
\hline
6 & \xmark & \xmark & \cmark & \cmark & M & L & 56.42 \\
\hline
7 & \xmark & \cmark & \xmark & \xmark & M & L & 54.75 \\
\hline
8 & \cmark & \cmark & \cmark & \cmark & M & L & 47.20 \\ 
\hline
9 & \cmark & \xmark & \xmark & \xmark & M & L & 44.83 \\ 
\hline
\end{tabular}
}
    \caption{Ablation study results on Flickr30k using PNASNet. SA: Softmax Attention; NLT: Non-Linear Text mapping; NLV: Non-Linear Visual mapping; WL: Word-Layer; SL: Sentence-Layer; Acc.: pointing game accuracy.}
    \label{tab:ablation}
\vspace{-0.1cm}
\end{table}

We first evaluated the efficacy of using multi-level feature maps (ML) with level selection compared to a fixed choice of visual layer (M: middle layer, L: last layer) for comparison to word and sentence embeddings (WL and SL).
Specifically, we used \textit{Cell~7} as middle layer, and \textit{Cell~11} as last layer, to be compared with word and sentence embedding in Eq.~(\ref{eq:heatmap_multi}) and Eq.~(\ref{rs}), respectively.
The results in rows $1,2$ show that using level-attention mechanism based on multi-level feature maps significantly improves the performance over single visual-textual feature comparison.

We then study the affect of non-linear mapping into the common space for the text and visual features (NLT and NLV). 
By comparing rows $2,4,5,7$, we see that non-linear mapping in our model is really important, and replacing any mapping with a linear one significantly degrades the performance.
We can also see that non-linear mapping seems more important on the visual side, but best results are obtained with both text and visual non-linear mappings.

We further study the use of ELMo for text embedding or the commonly used approach of training a Bi-LSTM. 
Specifically, we simply replaced the pre-trained BiLSTMs of ELMo model with a trainable BiLSTM (on top of word embeddings of ELMo), and directly feed the BiLSTM outputs to the attention model. 
The results in rows $1,3$ and $2,6$ show the importance of using a strong contextualized text embedding as the performance drops significantly.

We also study the use of softmax on the heatmaps, comparing rows $2,8$, we see that applying softmax leads to a very negative effect on the performance.
This makes sense, as elaborated in Section~\ref{sec:mlmm_attention}, since this commonly used approach forces unnecessarily the heatmap to have a distribution on either words or regions. 
Results in row $9$ corresponds to a simple baseline on par with the state-of-the-art, showing how much improvement can be gained by not using softmax, the use of our multi-level non-linear common space representation and attention mechanism, and a powerful contextualized textual embedding.

\begin{figure}[tb]
    \centering
    \includegraphics[width=\columnwidth]{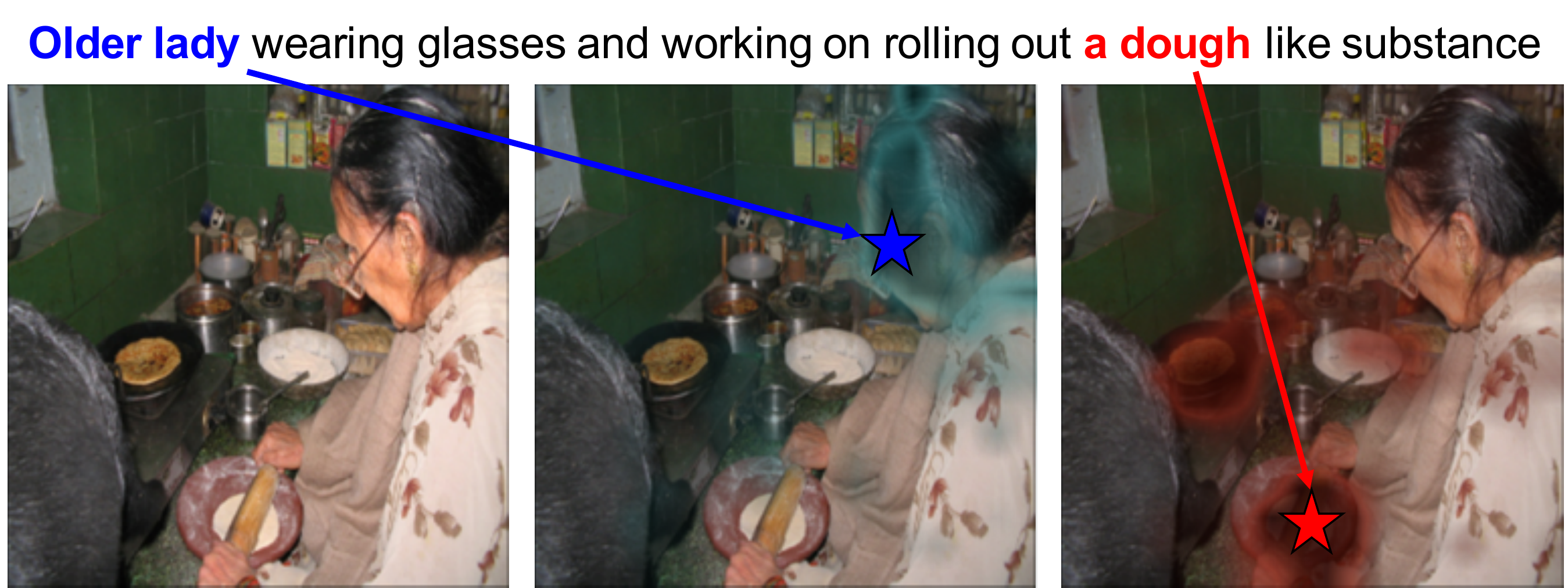}\\
    \includegraphics[width=\columnwidth]{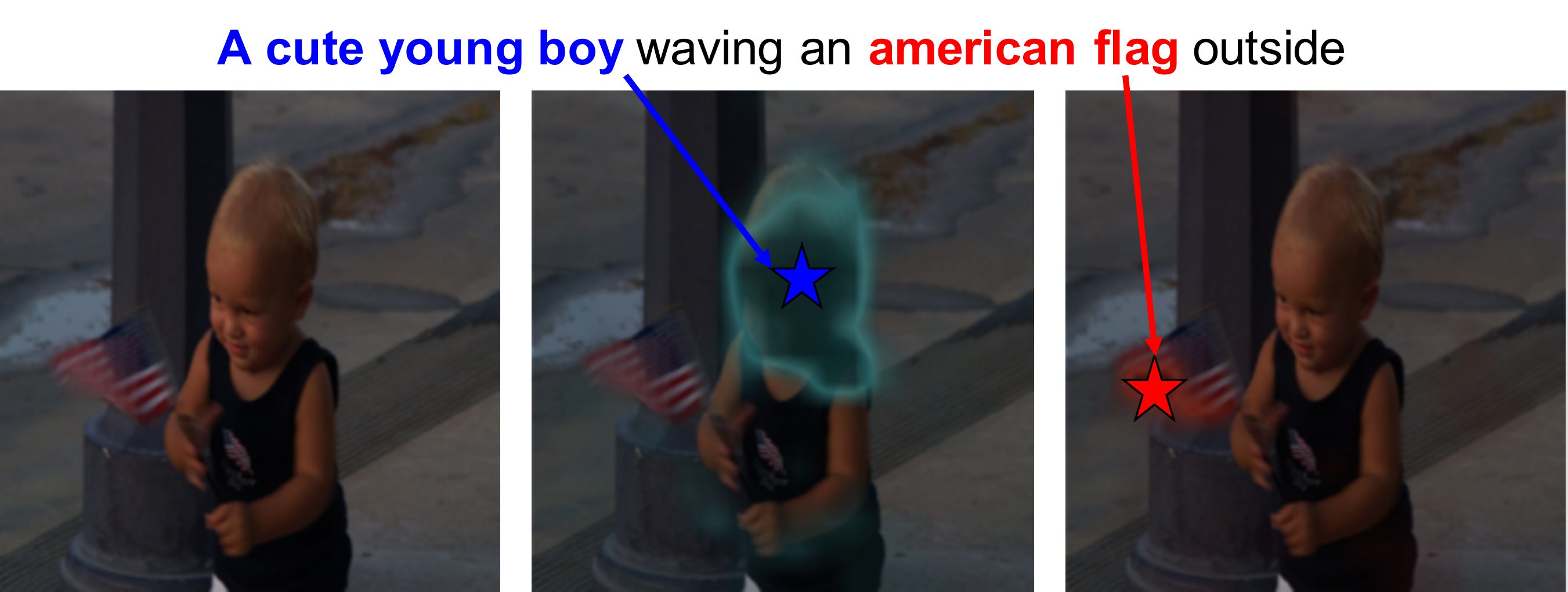}\\
    \includegraphics[width=\columnwidth]{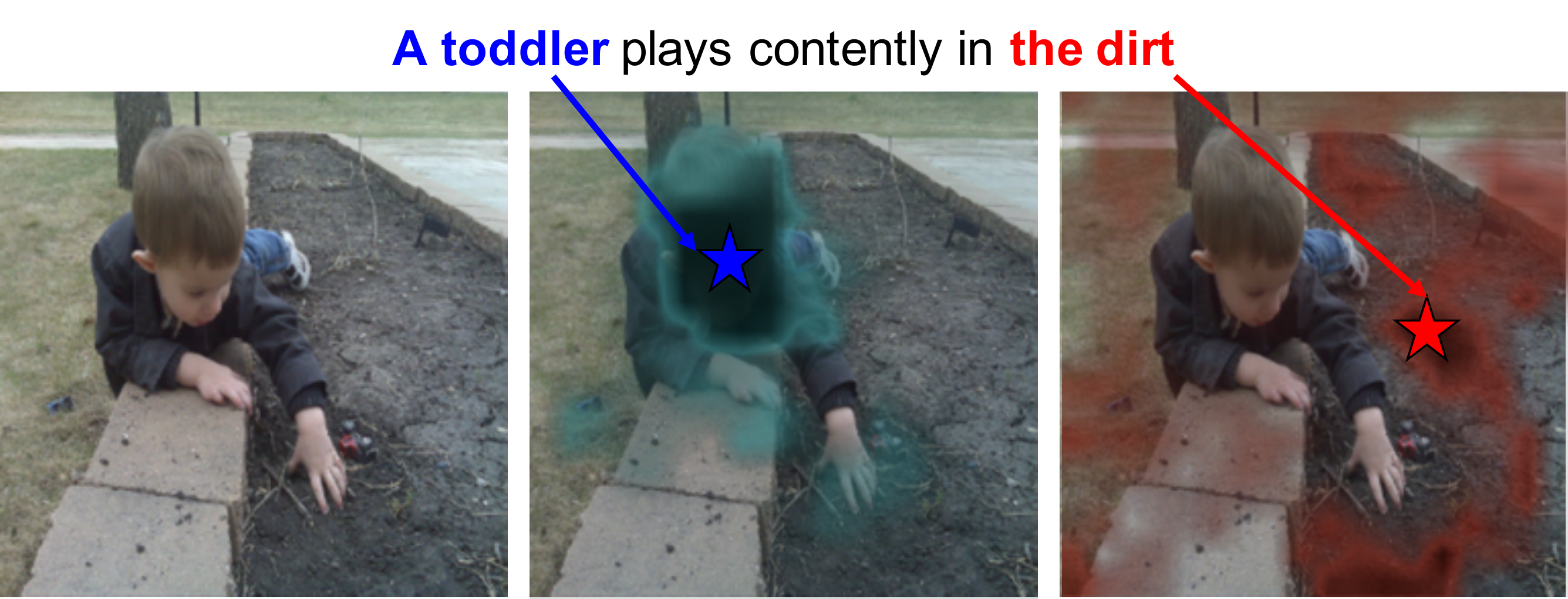}\\
    \includegraphics[width=\columnwidth]{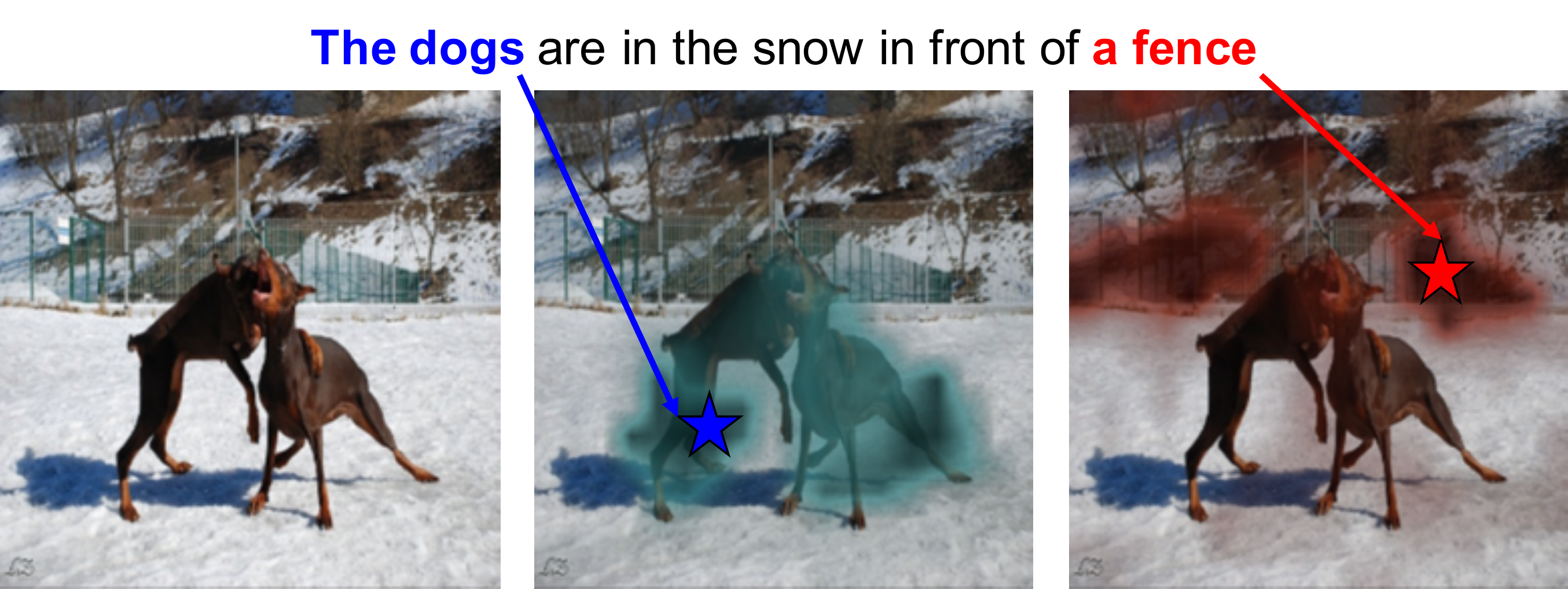}\\
    \includegraphics[width=\columnwidth]{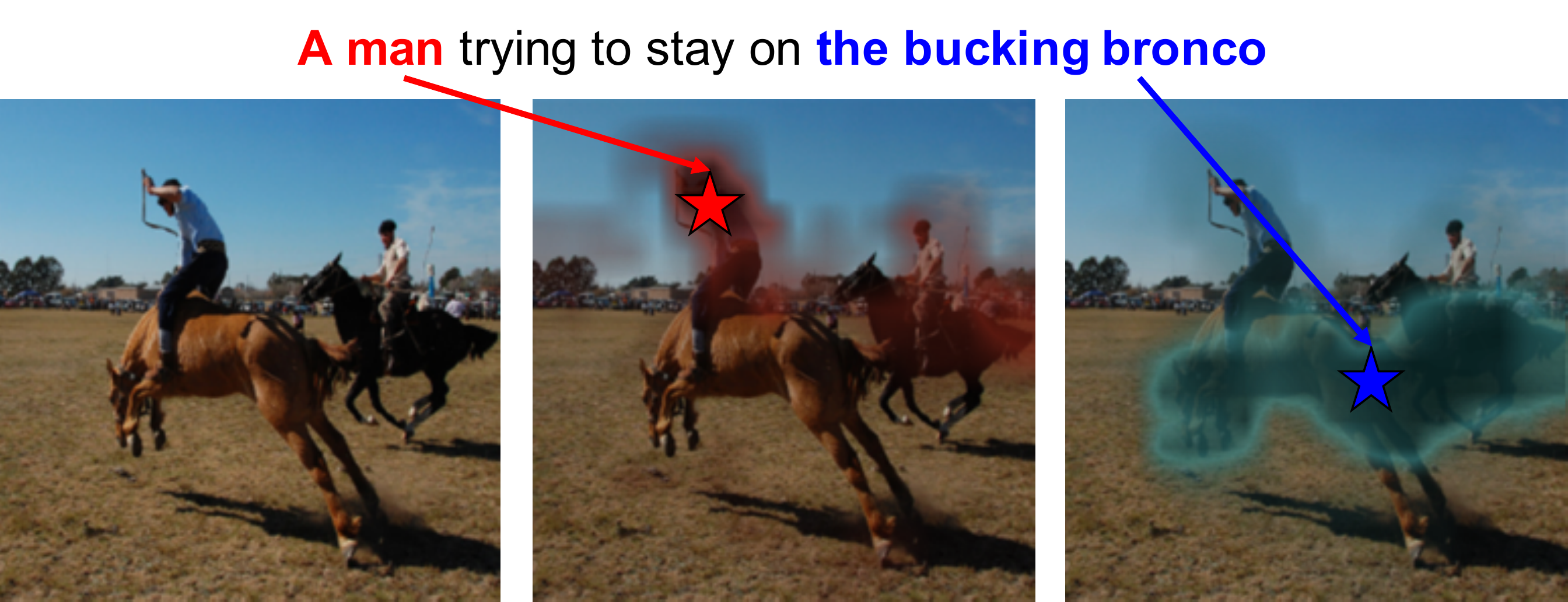}\\
    \caption{Some image-sentence pairs from Flickr30k, with two queries (colored text) and corresponding heatmaps and selected max value (stars).}
    \label{fig:qualitative_flickr30k}
    \vspace{-0.15cm}
\end{figure}

\begin{figure}[tb]
    \centering
    \includegraphics[width=\columnwidth]{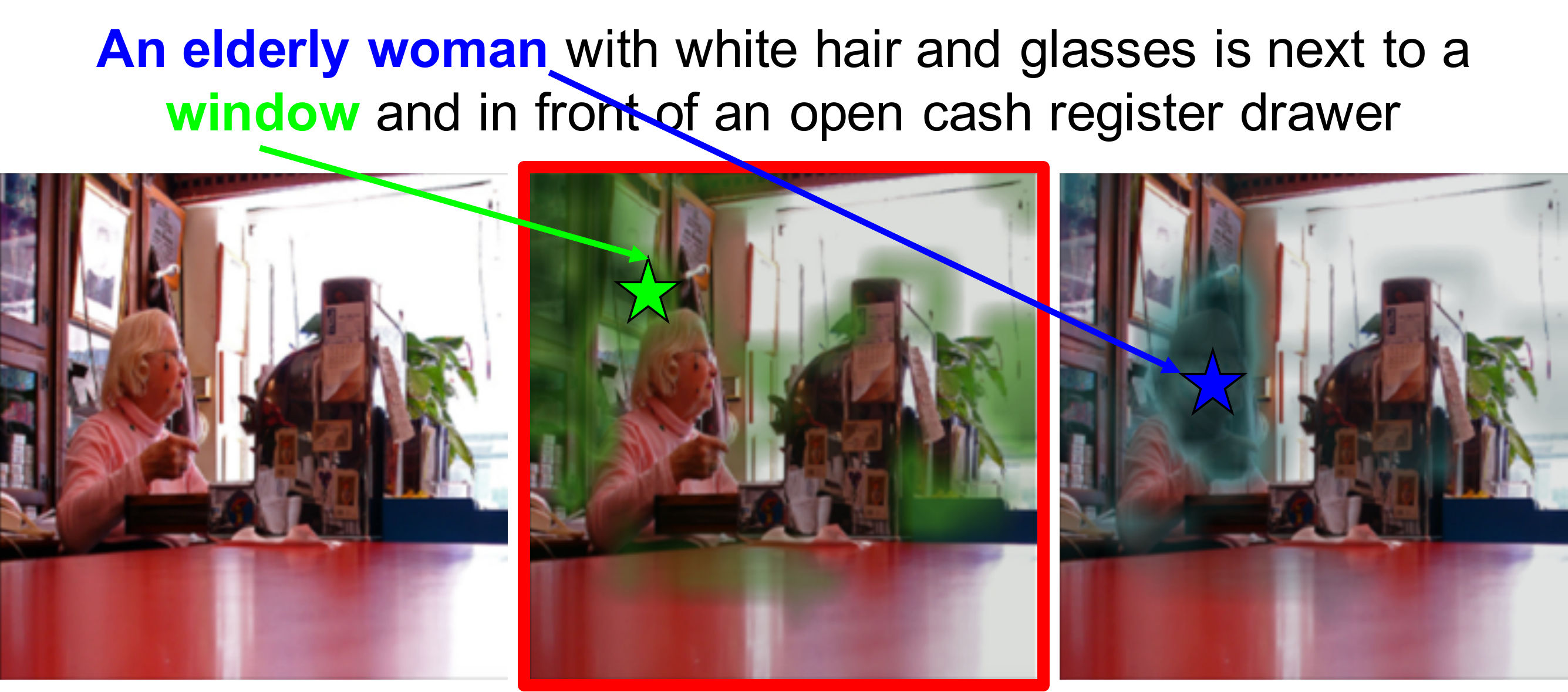}\\
    \includegraphics[width=\columnwidth]{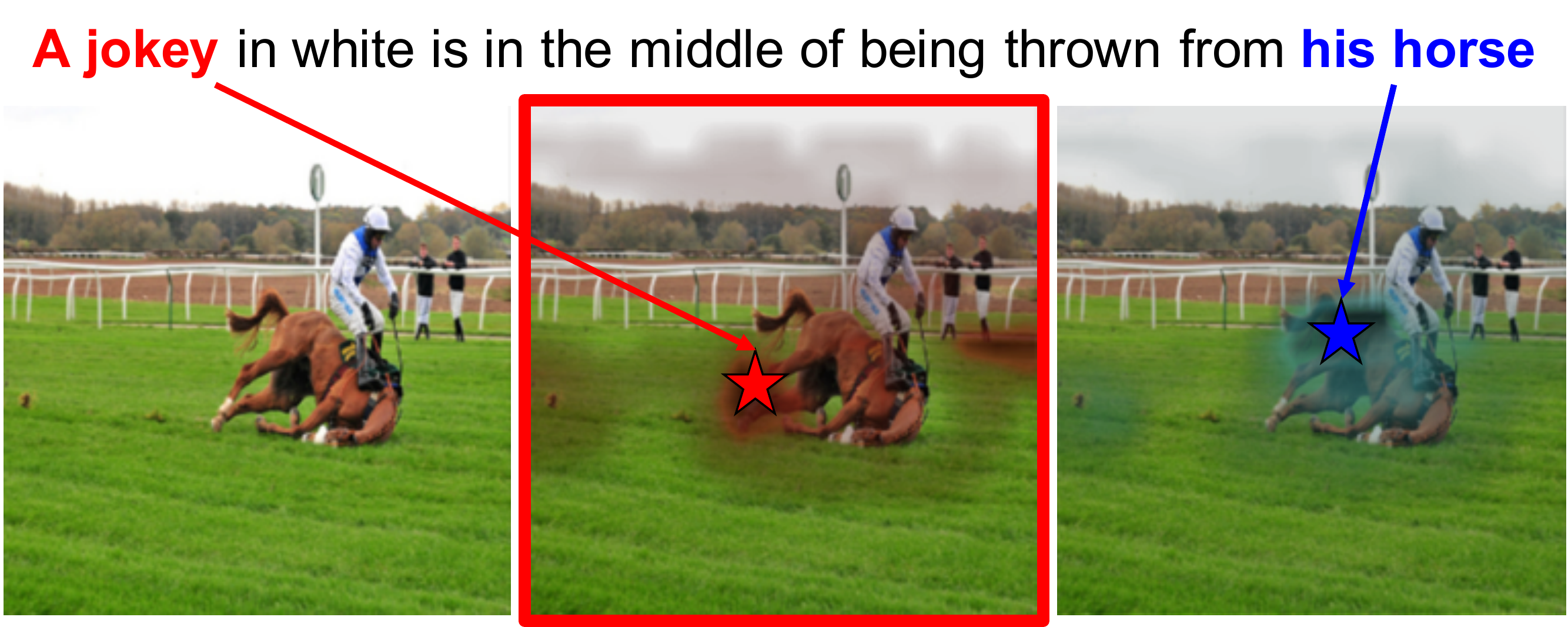}\\
    \caption{Some failure cases of our model. 
    The model makes some semantically reasonable mistakes in pointing to regions.
    }
    \label{fig:qual_bad}
    \vspace{-0.25cm}
\end{figure}

\subsection{Qualitative results}
We give in Figure~\ref{fig:qualitative_flickr30k_big}, \ref{fig:qualitative_flickr30k}, and~\ref{fig:qual_bad} some examples of heat maps generated for some queries of the Flickr30k dataset.
Specifically, we upsample the heatmaps from their original size of $18 \times 18$ (as we use the VGG backbone for these visualizations) by bilinear interpolation to the original image size. 
We can observe that the max (pointing) location in heatmaps point to correct location in the image and the heatmaps often capture relevant part of the image for each query. 
It can deal with persons, context and objects even if they are described with some very specific words (e.g. "bronco"), which shows the power of using a character-based contextualized text embedding. 
Finally, Figure~\ref{fig:qual_bad} shows some localization failures involving concepts that are semantically close, and in challenging capture conditions. 
For example, the frames are mistakenly pointed for the query "window" which is overexposed.

\section{Conclusion}
In this paper, we present a weakly supervised method for phrase localization which relies on multi-level attention mechanism on top of multi-level visual semantic features and contextualized text embeddings. 
We non-linearly map both contextualized text embeddings and multi-level visual semantic features to a common space and calculate a multi-level attention map for choosing the best representative visual semantic level for the text and each word in it.
We show that such combination sets a new state of the art performance and provide quantitative numbers to show the importance of 1. using correct common space mapping, 2. strong contextualized text embeddings, 3. freedom of each word to choose correct visual semantic level.
Future works 
lies in studying 
other applications such as Visual Question Answering, Image Captioning, etc.

\section*{Acknowledgment}
This work was supported by the U.S. DARPA AIDA Program No. FA8750-18-2-0014. The views and conclusions contained in this document are those of the authors and should not be interpreted as representing the official policies, either expressed or implied, of the U.S. Government. The U.S. Government is authorized to reproduce and distribute reprints for Government purposes notwithstanding any copyright notation here on.

{\small
\bibliographystyle{ieee}
\bibliography{egbib}
}

\end{document}